\pdfoutput=1
\documentclass[11pt]{article}

\usepackage{acl}

\usepackage{times}
\usepackage{latexsym}

\usepackage[T1]{fontenc}
\usepackage[utf8]{inputenc}

\usepackage{microtype}

\title{KUL@SMM4H’23: Text Augmentations with R-drop for Classification of Tweets Self Reporting Covid-19}

 \author{Sumam Francis \and Marie-Francine Moens\\
        KU Leuven, Belgium }
\begin{document}
\maketitle
\begin{abstract}

This paper presents models created for the Social Media Mining for Health 2023 shared task. Our team addressed the first task, classifying tweets that self-report Covid-19 diagnosis. Our approach involves a classification model that incorporates diverse textual augmentations and utilizes R-drop to augment data and mitigate overfitting, boosting model efficacy. Our leading model, enhanced with R-drop and augmentations like synonym substitution, reserved words, and back translations, outperforms the task mean and median scores. Our system achieves an impressive F1 score of $0.877$ on the test set.

\end{abstract}

\section{Introduction}

The goal of the shared task as part of the  Social Media Mining for Health (SMM4H-23) \cite{SMM4H2023}\footnote{This paper has been peer-reviewed and accepted for presentation at SMM4H'23 at AMIA 2023 Annual Symposium} is to facilitate the use of Twitter data for monitoring personal experiences of Covid-19 in real-time and on a large scale. The task involves automatically distinguishing tweets that self-report a Covid-19 diagnosis, for example, a positive test, clinical diagnosis, or hospitalization from tweets that merely state that the user has experienced Covid-19  without presenting any evidence and thus would not be considered a diagnosis. 

SMM4H Task 1 comprises datasets of English tweets with annotations indicating the presence or absence of Covid-19 diagnosis in the tweet. Recent research has highlighted the benefits of data augmentation, as it enhances the diversity  of training data and  improves the robustness of the model on downstream tasks. The majority of data augmentation techniques generate new examples through modifications of existing ones, rooted in previous task-specific knowledge. Augmented data functions as a form of regularization, for instance, R-drop \cite{rdrop}, effectively mitigating overfitting during the training of machine learning models.

\section{Data}
%\subsection{Dataset}

The dataset \cite{data} comprises a training set ($7600$ tweets), validation set ($400$ tweets), and test set ($10,000$ tweets). The dataset is imbalanced  with only  around $20\%$ of the tweets consisting of Covid-19 diagnosis mentions. We use oversampling to deal with this class imbalance.
The evaluation metric used is the  F1-score for the “positive” class (i.e., tweets that self-report a Covid-19 diagnosis). The training data include the Tweet ID, the tweet text, and the annotated binary label.
%\subsection{Pre-processing}

As part of the pre-processing on the dataset, we removed URLs, retweets, mentions, extra space, non-ascii words, and characters. Further, we lower-cased and striped off white spaces at both ends. We inserted space between punctuation marks. 

\section{Model}

For task 1, we combine two strategies: 1) textual augmentations to enhance training data diversity, and 2) fine-tuning with pretrained transformer models. Our approach uses a  Covid-Twitter-BERT (CT-BERT \cite{Covidtwitterbert}), which  is trained on 97M unique tweets (1.2B examples) related to Covid-19 exhibiting superior performance in tweet-related NLP tasks. Further, to mitigate the effects of an imbalanced label distribution and reduce overfitting, we employed the following strategies to enhance the model performance.

\textbf{R-drop} \cite{rdrop}:  This technique is a simple yet highly effective regularization method built upon dropout. It minimizes the bidirectional Kullback-Leibler (KL) divergence between output distributions of dropout-generated sub-model pairs during training. Input data undergoes dual forward passes, producing distinct prediction distributions. R-drop enforces regularization by minimizing the bidirectional KL divergence between these distributions for the same input, aiding data augmentation and mitigating overfitting.

The different textual augmentations used include the following. \textbf{Synonym substitution}: This augmentation involves the random replacement of words in the tweet with their synonymous WordNet counterparts. \textbf{Tense transformation}: This conversion alters the tense of English sentences, e.g., from present to past. \textbf{Reserved tokens replacement}: Tokens within tweets were substituted using predefined reserved tokens. For instance, the Covid-19 tokens were subject to a random replacement from a reserved list of alternatives like "Coronavirus," "Covid," "corona," and "SARS-CoV-2." etc. \textbf{Back translation}: This transformation translates an input sentence from the source language to the target language and subsequently translates it back to the source language. We use the English-German translation model utilizing it as a method for nuanced paraphrasing.

\begin{table}[h!]
  \caption{Micro-average Precision (P), Recall (R) and F1 scores (F1) on the validation set of the SMM4H 2023 Task 1 with CT-BERT model.}
  \label{tab:val_results}
  \addtolength{\tabcolsep}{-1.4mm}
  \centering
  \begin{tabular}{lccl}
    \hline
    Augmentation   & P & R & F1\\
    \hline
    - &  0.8438 &0.9373 & 0.8925 \\
    \hline
    + Text-aug &   0.8593& 0.9649&0.9090\\
    + Text-aug + R-drop& \textbf{0.9003} &\textbf{0.9473} &\textbf{0.9230}\\
    \hline

  \end{tabular}
\end{table}

\begin{table}[h!]
  \caption{Micro-average Precision (P), Recall (R) and F1 scores (F1) on the test set of the SMM4H 2023 Task 1 with CT-BERT model.}
  \label{tab:test_results}
  \addtolength{\tabcolsep}{-0.75mm}
  \centering
  \begin{tabular}{lccl}
    \hline
     Augmentation & P & R & F1\\
    \hline

    + Text-aug &0.859 &0.869 & 0.864\\
    
    + Text-aug + R-drop & \textbf{0.888} & \textbf{0.866} & \textbf{0.877}\\
    \hline

    Task mean results   & 0.824 &0.792  &	0.804\\
    Task median results &0.853 &0.861 &0.865\\
    \hline
    
  \end{tabular}
\end{table}

\section{Experiments and Results}

For classification, each model is fine-tuned over $10$ epochs with a learning rate of $5e-5$ using Adam optimizer. Batch size is set to $32$, and max sequence length to $128$. We use PyTorch and HuggingFace \footnote{https://huggingface.co/models} for CT-BERT training, applying cross-entropy loss. Model checkpoints are saved every $200$ steps based on the validation set's F1-score.
We adapted CT-BERT's loss function, adding KL divergence for (R-drop) regularization. Training data was enriched with textual augmentations from section $3$ (Text-aug).

Incorporating R-drop and textual augmentations into CT-BERT produces the highest accuracy for detecting Covid-19 diagnosis in tweets (see Table~\ref{tab:val_results} and Table~\ref{tab:test_results}). Augmentations diversify data, boosting model robustness and generalization. Noise from augmented data reduces overfitting risk and the model's reliance on specific patterns. Augmented examples aid learning and capture linguistic nuances for better language understanding. Adding R-drop further enhances performance, akin to an ensemble of sub-models within a single model. These sub-models encourage the model to capture various aspects of the data and promote learning more generalized features. Results surpass the shared task's mean and median scores, showcasing the effectiveness of this approach.

\section{Conclusion}
In this work, we developed a classification model enhanced with R-drop and textual augmentations to mitigate label imbalance and avoid the risk of overfitting thereby improving model performance.  We demonstrate the significance of the textual augmentations and R-drop in enhancing model generalizability. Our system achieves an impressive F1 score of $0.877$ on the test set.

\bibliography{anthology,custom}
\bibliographystyle{acl_natbib}

\end{document}